\documentclass[journal]{IEEEtran}

\usepackage{fancyhdr} 

\fancypagestyle{plain}{
  \fancyhf{}
  \fancyhead[C]{Conference on \LaTeX}     %% C or L or R.
  \fancyfoot[L]{This is a notice}%                        %% C or L or R.

}
\usepackage{eso-pic}

\usepackage{cite}
\usepackage{svg}
\usepackage{soul,color}
\usepackage{amsmath,amssymb,amsfonts}
\usepackage{graphicx}
\usepackage{textcomp}
\usepackage{xcolor}
\usepackage{booktabs}
\usepackage{comment}
\usepackage{hyperref}
\usepackage{amsmath,amsfonts}
\usepackage{algorithmic}
\usepackage{graphicx}
\usepackage{textcomp}
\usepackage{xcolor}
\usepackage{enumitem}
\usepackage{hyperref}
\usepackage{balance} 
\usepackage{multirow}
\usepackage{color,soul}

\def\BibTeX{{\rm B\kern-.05em{\sc i\kern-.025em b}\kern-.08em
    T\kern-.1667em\lower.7ex\hbox{E}\kern-.125emX}}
\DeclareUnicodeCharacter{2730}{}
\ifCLASSINFOpdf
\else
\fi
\begin{document}
\title{Past, Present, and Future: A Survey of The Evolution of Affective Robotics For Well-being}

\author{Micol Spitale$^{1,2}$,~\IEEEmembership{}
        Minja Axelsson$^{1}$,~\IEEEmembership{}
        Sooyeon Jeong$^{3}$, ~\IEEEmembership{}
        Paige Tutt\"os\'i$^{4}$,~\IEEEmembership{}
        Caitlin A. Stamatis$^{5}$,~\IEEEmembership{}
        Guy Laban$^{1}$,~\IEEEmembership{}
        Angelica Lim$^{4}$,~\IEEEmembership{}
        Hatice Gunes$^{1}$~\IEEEmembership{}
        % <-this % stops a space
\thanks{$^1$ University of Cambridge, UK; $^2$ Politecnico di Milano, Italy, $^3$ Purdue University, USA, $^4$ Simon Fraser University, Canada, $^5$ Northwestern University, USA; contact author email: micol.spitale@polimi.it}% <-this % stops a space
}

%https://www.overleaf.com/project/65686a19627074f2411bd3ab

\markboth{Journal of \LaTeX\ Class Files,~Vol.~14, No.~8, August~2015}%
{Shell \MakeLowercase{\textit{et al.}}: Bare Demo of IEEEtran.cls for IEEE Journals}



\maketitle

\begin{abstract}
%\hl{@Minja: TODO} 

Recent research in affective robots has recognized their potential in supporting human well-being. Due to rapidly developing affective and artificial intelligence technologies, this field of research has undergone explosive expansion and advancement in recent years. In order to develop a deeper understanding of recent advancements, we present a systematic review of the past 10 years of research in affective robotics for wellbeing. In this review, we identify the domains of well-being that have been studied, the methods used to investigate affective robots for well-being, and how these have evolved over time. We also examine the evolution of the multifaceted research topic from three lenses: technical, design, and ethical. Finally, we discuss future opportunities for research based on the gaps we have identified in our review -- proposing pathways to take affective robotics from the past and present to the future. The results of our review are of interest to human-robot interaction and affective computing researchers, as well as clinicians and well-being professionals who may wish to examine and incorporate affective robotics in their practices.

%Developing affective robots for well-being requires advancement in software and hardware technology, as well as in special considerations for the vulnerable target populations, cooperation with health experts, as well as design and ethical considerations. Despite this multi-disciplinary nature, no previous studies have provided a comprehensive snapshot of the last decade in affective robotics for well-being in terms of  technical, design and ethical aspects. 
%This survey reviews how the studies of affective robotics for well-being have evolved in their technology, design considerations, and ethical investigations over the last decade (from 2013 to 2022). We highlighted the trajectory of the research in this field, by identifying past and present gaps and future opportunities in affective robotics. 
\end{abstract}


% \fi
\IEEEpeerreviewmaketitle

\section{Introduction}
\label{sec:introd}
%\hl{@sooyeon: Hatice suggested to restructure completely the introduction. you can take a look to her comments directly. please let me know if you cannot do this.}

According to the World Health Organization (WHO), well-being is defined as ``a positive state experienced by individuals and societies'' and ``is a resource for daily life and is determined by social, economic and environmental conditions \cite{nutbeam2021health}''. Yet, approximately 1 in every 8 people, or 970 million individuals worldwide, were living with a mental disorder in 2022, and mental, neurological, and substance use disorders account for 10\% of the global burden of disease and 25.1\% of non-fatal disease burden \cite{gray2022mental,institute2021global,who2022}, and more than 80\% of people with mental health conditions lack access to quality and affordable care \cite{world2019special}. This has created a pressing need to support people's well-being. 

Affective Robotics offers a promising approach to enhance human-robot interaction and improve overall well-being by studying how robots can recognize, interpret, process, and simulate human affect \cite{spitale2022affective}. Since the beginning of research in social robotics  \cite{breazeal2003toward} in the early 2000's, the ability to understand human affect and emotion has played a key role in enabling robots to be helpful in various application areas \cite{kennedy2016social,gonzalez2011maggie,robinson2021humanoid}, especially for mental health and well-being support \cite{kabacinska2021socially,robinson2021humanoid, rhim2019investigating}. % to improve human lives by leveraging the advancements in the affective computing field. 
For example, affective robotics have been used to lead mindfulness meditations \cite{axelsson2022robots}, facilitate social bonding \cite{rhim2019investigating}, support diet and physical activity \cite{robinson2021humanoid}. 
In these contexts, being able to recognize and generate affect is extremely important because the goal of the interaction involved tracking or supporting a ``positive state" in the user.

However, studying, developing, and designing affective robots for well-being is still an open research area due to multiple challenges in this research landscape.
First, affective robotics is an interdisciplinary research topic that combines the fields of Affective Computing and Human-Robot Interaction. This intersectionality brings together multiple disciplines, making it challenging to consider various aspects simultaneously (\textbf{Challenge 1}, C1). To address this complexity, the field of affective robotics must tackle several challenges. It should develop computational models that accurately understand, model, and adapt to human behaviors. These models must be embedded into robotic platforms that can be used in real-world applications, such as well-being (\textit{technical} challenge). Affective robots must be designed with features that enable smooth and seamless interaction with humans. This includes understanding how to design the robots themselves and how they can effectively interact with humans (\textit{design} challenge). Affective robotics has a responsibility to develop fair and ethical computational models, particularly in high-stake application scenarios like well-being. Additionally, the field must conduct studies that are ethically compliant and ensure the well-being of both humans and robots involved in these interactions (\textit{ethics} challenge).

Second, affective robotics relies on advancements in affective computing, which utilizes cutting-edge artificial intelligence (AI) models to understand human emotional states. However, the field of AI has undergone \textit{rapid evolution}, particularly in recent years, with an unprecedented growth in AI advancements (\textbf{Challenge 2}, C2).

%Third, research on affective robots for well-being has primarily focused on a few sub-areas. A recent review conducted by Guemghar et al. \cite{guemghar2022social} reports that the majority of social robot studies related to well-being were applied in the domain of dementia, while there were only a few studies conducted in other types of cognitive impairment, such as schizophrenia, depression, ADHD, and intellectual disability. We believe affective robots have opportunities to support patients and extend clinical care systems across diverse well-being application domains. Recently, there has been more effort in the field of HRI to collaborate with domain experts, such as mental health professionals and well-being experts, and to include stakeholders in the design process and research process, e.g. \cite{rebola2021co, bjorling2019participatory}. However, it is still unclear how much these key stakeholders are involved in the research process and how these interdisciplinary collaborations are conducted (\textbf{Challenge 3}, C3). .
Third, research on affective robots for well-being has primarily focused on a few sub-areas, such as dementia, with limited studies in other cognitive impairments like schizophrenia, depression, ADHD, and intellectual disability. Recent efforts involve collaborating with domain experts, such as mental health professionals and well-being experts, and including stakeholders in the design and research processes. However, it remains unclear how affective robotics can support patients and extend clinical care systems across diverse well-being domains and how much these stakeholders are involved in the research process and how interdisciplinary collaborations are conducted (\textbf{Challenge 3}, C3).

Therefore, we conducted a literature review of the last 10 years of research in affective robotics for well-being from different point of views, namely technical, design and ethical, to better understand the multi-facet complexity of this research field (addressing C1, \textbf{Solution 1}). This review focused on the trajectory of the affective robots for well-being. Specifically, we identified existing gaps and areas that need further investigation, and provided insights into future opportunities to improve the state of the art and close the gaps (addressing C2, \textbf{Solution 2}). In this review, we have also identified: i) domains of well-being that have been studied in affective robotics research and establish which areas are still unexplored and warrant further exploration, ii) the current methods used to investigate affective robotics for well-being, and iii) how these research processes and methodologies have changed over time (addressing C3, \textbf{Solution 3}). 

\section{Background and Definitions}
\label{sec:back}

%\begin{itemize}
%    \item Affective robotics: intersection between human-robot interaction and affective computing
%    \item Applications scenarios of affective robots
%    \item Current systematic reviews on this topic
%    \item How we are contributing to the current state of the art, and how our survey can help other researchers
%\end{itemize}
The field of Affective Robotics (AR) has been increasingly applied in various domains, such as healthcare and well-being applications, where affective robots have demonstrated significant potential \cite{gunes2023affective}. 
With the term well-being, we refer to a comprehensive concept that encompasses what it means to be functioning as a healthy person across multiple domains \cite{gellman2020encyclopedia}. For example, mental well-being is about achieving a positive state of mind  \cite{warr1990measurement} and it involves aspects like the ability to cope with challenges, recognizing personal strengths, and finding purpose in daily life.

Affective Robotics is at the intersection between Affective Computing (AC) and Human-Robot Interaction (HRI) fields. AC is an emerging interdisciplinary field that integrates the affective and computational sciences and studies how machines can measure human affective states \cite{picard2000affective}. Analogously, HRI is also an interdisciplinary field that encompasses expertise from cognitive, psychology, robotics, design  and computer sciences, and it aims at understanding the dynamics in human-robot interactions \cite{spitale2023systematic}. 
As a result, Affective Robotics inherited the interdisciplinary nature of both AC and HRI fields when applied in a well-being context. From a computational and \textbf{technical} point of view \cite{spitale2022affective}, affective robotics research needs to: (i) understand human behaviours while interacting with robots; (ii) model human-inspired behaviours in robots; (iii) adapt to the interaction context to meet the personal needs of humans interacting with the robot; and (iv) translate the results obtained in controlled settings into real-world scenarios without compromising performance and efficiency.
From a \textbf{design} point of view (i.e., how to design robot features and interactions with humans), the affective robotics field lags behind the advances in the HRI field in which researchers collaborate with domain experts, e.g., teachers, psychologists, to design robots that can be used in real-world scenarios \cite{axelsson2022robots}. The involvement of stakeholders is not an easy task, but it becomes fundamental when applying technologies such as robots to high stake  contexts like well-being. 
From an \textbf{ethical} perspective (i.e., considering moral, value and legal implications), the potential for both positive and negative outcomes in well-being context makes it an ethically complex issue, requiring careful ethical consideration to achieve a balance between the positive and negative aspects \cite{devillers2023ethical}. The AC community has made efforts in this direction by including a mandatory Ethical Statement section in the International Conference on Affective Computing \& Intelligent Interaction (ACII) submissions, and by investigating the current ethical issues, positive and negative, which arise from the current state of affective computing \cite{devillers2023ethical}. However, these efforts have not yet been translated into the area of affective robotics.

Previous efforts have been made to survey the current state of the art of affective robotics field, e.g., \cite{spitale2022affective, olugbade2023touch, scoglio2019use,robohealth2024}. For example, \cite{spitale2022affective} reviewed the last decade of affective robotic works in well-being focusing only on technical aspects without including any considerations for ethics or design. \cite{olugbade2023touch} conducted a literature review on affective touch during human-agent and human-robot interactions. \cite{scoglio2019use} reviewed the past works on robotics for mental health and well-being without focusing on affective aspects, similarly to \cite{robohealth2024} which focused exclusively on the introduction of robots in health psychology applications (e.g., behavioural change and emotion regulation interventions). 
None of these previous surveys have provided a comprehensive snapshot of the last decade in affective robotics for well-being and a future research agenda for the field focussing on the multi-disciplinary aspects that characterise it, namely technical, design and ethical.

Therefore, in this paper, we conducted a survey to review the papers from the last decade on affective robotics for well-being by analysing the evolution in this research field from technical, design and ethical point of views.

%Past works have provided different definitions \cite{}, for example \cite{} defines affective robots as "..", but overall literature 

\section{Method}
%\subsection{Terminology}
% define the terminology we are using in reporting

This section describes the methodology to identify the papers included in this survey by reporting the procedure, the search query, the inclusion criteria, the selection process, the data analysis and extraction, and the terminology used.

\subsection{Procedure}
\begin{figure}
    \centering
    \includegraphics[width = \columnwidth]{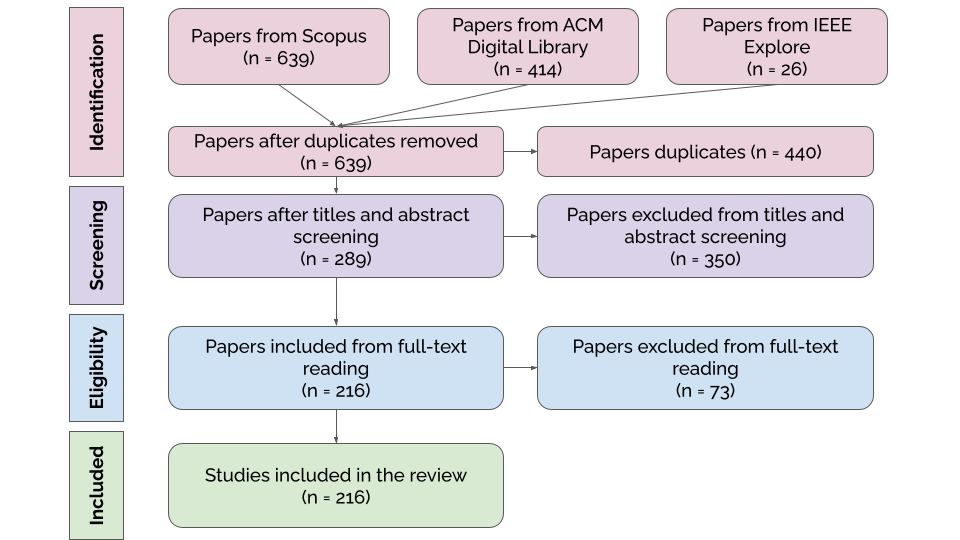}
    \caption{PRIMA Schema for our systematic review.}
    \label{fig:prisma}
\end{figure}
% overall survey procedure from literature screening to analysis
To define our systematic literature review approach, we followed the guidelines established by Nightingale \cite{nightingale2009guide} and we expanded upon a previous preliminary survey on this topic undertaken by two authors of this work \cite{spitale2022affective}. Our methodology adhered to the Preferred Reporting Items for Systematic Reviews and Meta-Analyses (PRISMA) framework \cite{moher2009preferred}, widely acknowledged as the gold standard for systematic reviews and meta-analyses. PRISMA ensures the quality and replicability of the review, organizes the final manuscript using standardized headings, and facilitates the assessment of the study's strengths and weaknesses \cite{nightingale2009guide}.
Specifically, the PRISMA framework presents an evidence-based, minimal set of reporting items for systematic reviews, structured around a four-phase flow depicted in Figure \ref{fig:prisma}. The initial phase involves identification, wherein all potential manuscripts are gathered. Subsequently, during the screening phase, papers meeting the eligibility criteria are selected based on an assessment of their titles and abstracts. In the selection phase, full texts are scrutinized, and only those meeting the same eligibility criteria are retained. Finally, in the inclusion phase, all selected papers undergo analysis to address the study's research questions.

%The following sections report the search query, our eligibility criteria, the selection process, disclose, and outline the methods employed for data extraction and analysis.

\subsection{Search Query}
We gathered papers based on the terms found in their titles, abstracts, and keywords, utilizing the Scopus, ACM Digital Library, and IEEE Xplore databases. Scopus was selected due to its broad coverage across multiple disciplines beyond computer science. The ACM Digital Library and IEEE Xplore were chosen for their extensive coverage in human-computer interaction, computer science, and engineering. The search queries were adjusted slightly to accommodate the requirements of each database. We provide the search query used for Scopus:
\begin{verbatim}
    TITLE-ABS-KEY ( ( "affective robotic*"  
    OR  "social robot*"  OR  
    "emotional robot*"  OR  "socially 
    assistive robot*" )  AND  ( 
    "wellbeing"  OR  "well-being"  OR
    "mental health"  OR  "health" ) )  
    AND  PUBYEAR  >  2012 AND 
    PUBYEAR <  2023
\end{verbatim}

%Following the execution of the search queries to identify potential papers for review, we manually supplemented the list with additional relevant papers based on our prior knowledge of the domain. 
We executed the search queries to identify potential papers for review. After that, we removed duplicate entries and stored references in a Google Sheet file.

\subsection{Eligibility Criteria}

Papers were included only if:
\begin{itemize}
    \item they address well-being or health (both physical and mental)
    \item they model, analyse, design, or discuss the ethics of affective capabilities of a robot 
    \item use/discuss physical robots (e.g., video-conference, video)
    \item their title, abstract, and keywords contain at least one keyword describing such technology and one keyword from the Search Query Keywords. 
    %\item they are surveys (i.e. review papers) that provide a contribution (e.g., tool or protocol);
    %\ recommendations, guidelines, checklist, protocol, design tool, taxonomy etc.
%Include the survey paper if in the abstract one of those contributions is reported, and if it is an uncertain case just mark it as M (maybe included) 
    \item the robot is acting as an agent not as a tool (medium, e.g., to connect people)
\end{itemize}

Papers were excluded if: 
\begin{itemize}
    \item they were published before May 2012 and after the day of actual running of the search query, i.e., Oct 31, 2022, 
    \item they are not in English, 
    \item they are not in peer-reviewed journals and conference proceedings,
    \item they are surveys, (not provide any additional contributions, e.g., tools or protocol),
    \item they are inaccessible to the authors,
    \item they are low quality (i.e., they do not report the details necessary to evaluate their eligibility)
\end{itemize}

\subsection{Selection Process}

All manuscripts collected from the databases were screened based on their titles and abstracts. Then, a thorough examination of the full texts was conducted to ensure compliance with the eligibility criteria.

Prior to the screening process, a random sample of 5 papers was selected from the 2022 Scopus search and evaluated by five reviewers to establish consensus on inclusion and exclusion criteria. All reviewers had a full agreement on the inclusion or exclusion of the sample papers.
The remaining papers were then randomly distributed among the reviewers, with each reviewer screening a subset individually. 
Screening of titles and abstracts was conducted according to the predefined eligibility criteria. In cases where a reviewer had uncertainty regarding a particular paper, it was flagged for further discussion among all reviewers. The same set of reviewers were subsequently assigned different subsets of manuscripts for full-text screening and data extraction.

In total, 639 papers were collected using the specified search query (refer to Figure \ref{fig:prisma}). These manuscripts underwent initial screening based on titles and abstracts, resulting in the selection of 289 papers. Their full texts were reviewed. This process yielded a final count of 216 manuscripts, as reported in the Supplementary Materials. Among these 216 manuscripts, a total of 239 studies were identified, with some papers containing multiple studies.

The comprehensive list of extracted papers is publicly accessible via a spreadsheet stored in the GitHub repository\footnote{https://github.com/Cambridge-AFAR/aff-rob-survey.git}.

\subsection{Data Analysis and Extraction}

We defined a set of variables that can help understand the evolution of this field -- encompassing various types such as numerical (e.g., number of participants in each study), categorical (e.g., type of the robot), or qualitative (e.g., design methods) -- by creating a codebook of variables that can be found in the same Github repository.
We extracted relevant information from each paper using an analytical approach to assign a value to each of the defined variables.
For numerical variables, we employed a descriptive statistics description, while qualitative variables were analyzed using pattern-based methods. Categorical variables were derived through either a top-down or bottom-up approach as in a previous work \cite{catania2023conversational}. The former involved pre-defining variable categories based on existing literature. The latter approach entailed extrapolating variables from the collected data among the selected papers.
These variables are collected in Table \ref{tab:variables}.

\begin{table}[]
    \caption{List of variables extracted for each paper, the corresponding type and example of values.}
    \label{tab:variables}
\footnotesize
    \centering
    \resizebox{\columnwidth}{!}
   {
    \begin{tabular}{lll}
            \toprule
\textbf{Variable}                                & \textbf{Type}        & \textbf{Example}        \\
      \midrule
Year                    & numerical & 2013, 2014, etc. \\
Authors' discipline                      & categorical & computer  science, psychology, etc. \\
Country                                & categorical & USA, Finland, Japan, etc.                      \\
Study type                                & categorical & qualitative, quantitative, etc.                     \\
Psychological target outcome            & categorical & anxiety, depression, etc.                     \\
Application scenario                    & categorical & education, therapy, play, etc.                \\
Experimental setting                    & categorical & school, lab, workplace, etc.             \\
Number of sessions                      & categorical   & single, less than 30,                                     \\
    &    &more than 30                                   \\
Number of participants                  & numerical   & 10, 22, etc                                  \\
Study design                            & categorical & single-session, within-subject, etc. \\
Target population                       & categorical & neurotypical, elderly, dementia, etc.           \\
Age range                               & numerical   & 7-10 years-old                             \\
Mean age                                & numerical   &  9.4, 75.3 etc  \\
Gender                                  & categorical & female, non-binary, male, others                                       \\
Robot model                             & categorical & NAO, Pepper, Mario, etc                       \\
Mode of interaction                      & categorical & physical, virtual, video, picture              \\
Robot operation                         & categorical & autonomous, wizard of Oz,   \\
                       &  & semi-autonomous      \\
Affective behaviour generation                         & qualitative & facial expressions, \\
&& expressive movements, etc.   \\
Generation autonomy                       & categorical & autonomous, wizard of Oz,   \\
                       &  & semi-autonomous      \\
Affective behaviour perception   & qualitative & human facial expressions, \\
&& human movements, etc.   \\
Perception autonomy                       & categorical & autonomous, wizard of Oz,   \\
                       &  & semi-autonomous    \\
Technological aim                         & categorical & Yes, No      \\
Design aim                         & categorical & Yes, No        \\
Ethical aim                         & categorical & Yes, No     \\
Inclusion of clinicians                       & categorical & Yes, No  \\
Theory-grounded                      & categorical & Yes, No                                        \\
Data collection method                       & categorical & questionnaires, survey etc.   \\
Computational models                       & categorical & statistics, ML, deep learning etc.  \\
Design approach                       & qualitative & participatory design, user-centered, etc.     \\
Stakeholder involvement                      & categorical & young adults, clinicians, etc.    \\
Ethics on user safety                    & qualitative & deception, human contact, etc.    \\
Ethics on societal impacts                    & qualitative & fairness, justice, bias, etc.     \\
Ethical implications                       & categorical & Yes, No    \\
Ethics approval (internal)                       & categorical & Yes, No    \\
Ethics approval (external)                       & categorical & Yes, No     \\
        \bottomrule
\end{tabular}
}

\end{table} 
\subsection{Terms}
The following paragraphs describe the terminology used in this survey. 

\paragraph{Author disciplines} We defined the authors' disciplines taking the categories defined by \cite{spitale2023systematic}. As a result, we identified the following 15 categories of disciplines: art, biology, business, communication, computer science, dentistry, design, education, engineering, humanities, literature, medicine, philosophy, psychology, and social sciences.

\paragraph{Study type} With this term, we refer to the type of study that was conducted by authors, specifically we categorised the studies into qualitative, quantitative, theoretical, and meta-analysis works.

\paragraph{Study session} With this term, we refer to the number of sessions in which participants of each study interacted with the affective robot.

\paragraph{Psychological target outcome} The psychological target outcome is the specific psychological construct that the robots target or treat \cite{Magyar2019},for example, anxiety, loneliness, depression, or well-being. Notably, some studies of affective robots focus on well-being, whereas others target specific mental health outcomes (e.g., depression; anxiety). Mental health is typically more narrowly defined as the absence of certain disorder-specific symptoms. Wellbeing represents a broader term that may encompass factors such as overall ability life satisfaction, ability to cope with stress, and resilience. 

\paragraph{Application context} This refers to the specific conditions and settings in which study and/or findings are intended to be applied, and we identified the following categories: educational (e.g, the robot teaches, population's learning), therapeutic (that includes both assessment and treatments or interventions \cite{catania2023conversational}), mental (i.e., improves psychological well-being, like stress, anxiety, can be corollary to main diagnosis), health (i.e., any paper which focuses on public health and general well-being contexts, and does not fit into either of the others), psychological (i.e., specific psychological phenomena, e.g., joint attention, theory of mind), home (i.e., robots intended to be used either at home or residential centers), creative (i.e., improve people creativity), play (i.e., robots that promote play and game activities), social (i.e., the development of a social relationship with the robot), others (i.e., if no context is specified). The definitions of such application contexts are in line with previous work by \cite{spitale2023systematic}.

\paragraph{Experimental setting} The experimental setting is the environment where participants interacted with the robot, for example laboratories, schools, homes, etc.

\paragraph{Study design} This is the design of the study that can be within-subjects, between-subjects, random control trails, among others \cite{bethel2007psychophysiological}. 

\paragraph{Mode of interaction} This variable refers to the modality of the interaction between the robot and the human. For example, if participants were asked to evaluate a robot just watching a video of the robot we labelled it as ``video" mode of interaction, while if the participants interacted physically with a robot that have been categorised as ``physical".

\paragraph{Robot operation} The robot operation refers to how the robot was controlled, and we identified the following categories: wizard-of-oz (if the robot was controlled by the researchers or participants in the study), semi-autonomous (following the definition in \cite{spitale2023systematic}, e.g., if the robot performed actions autonomously but the decision are controlled by the researchers), and autonomous (if the robot could perform autonomously the whole interaction).

\paragraph{Affective behaviour generation and perception} With these variables, we want to label studies according to the capabilities of the robot to generate or express affect-based behaviours (e.g., facial expressions), and the capabilities of the robot to perceive affect-based behaviours following the suggestions in \cite{spitale2022affective}.

\paragraph{Aims} Motivated by the multi-disciplinary nature of the affective robotics field and the challenges that emerged in the literature (see Section \ref{sec:back}), we defined three specific aims for the studies surveyed: technical (i.e., studies that  examine AR from a technological perspective, for example works that presented a computational model to automatically detect emotions in human-robot interaction), design (i.e., studies that  examine AR from a design perspective, for examples works that presented a focus group to design features of an affective robot), and ethical  (i.e., studies that  examine AR from an ethical perspective, for example works that have investigated the ethical implications of introducing robots in public spaces).

\paragraph{Theory grounded} This variable refers to the studies that used and implemented systems, artifacts, or features backing it up with psychological theories (e.g., theory of mind).

\paragraph{Data collection method} This variable refers to the  methodology adopted for collecting data in studies, for example questionnaires, interviews, and surveys.

\paragraph{Computational models} This variable refers to the type of computational models used in the studies to develop the affective system/artifact/features. 

\paragraph{Design approach and stakeholder involvement} These variables include which design methods have been used to design AR, the interaction, or parts of the robot and whether stakeholders were involved or not in the design process of such AR. For eample, we categorized surveyed papers by the following criteria: 1) if the paper claimed to apply participatory or co-design, we assigned it to that label, 2) if the paper claimed neither participatory or co-design, but described the involvement of users, we assigned it a user-centred design label, 3) if the paper claimed neither of those, we assigned it as Not Reported.

\paragraph{Ethics on user safety, society, and implications} These variables encompass whether the paper discussed implications of user safety (e.g., deception, human contact) or  societal impacts (e.g., fairness, bias, and justice). Moreover, notes were collected on the specific aspects and categories of ethical discussion that were later organised and classified based on the framework in \cite{Riek2014ACO}. 

\paragraph{Ethics approval} With this variable, we labelled all the studies that have received an approval from internal (e.g., university) and/or external (e.g., hospital) entities.

\section{Evolution of Affective Robots for Well-being}
\label{sec:evolution}

This systematic review aims to better understand the evolution of affective robots for well-being in the last decade, from technological, design, and ethical perspectives to tackles the challenges C1 and C2 reported in Section \ref{sec:introd}. %\hl{@MICOL: why? once the introduction has been refined i can add the motivation here as well in line with what we have stated there}
This section describes the evolution of technological, design, and ethical aspects in the context of affective robotics.

\subsection{Technical and Computational Advancements Over Time}

We found 21 studies that have only a technical contribution (i.e., their main contribution was focusing only on technical aims) among 81 studies that also focused on technical aspects alongside design and ethical aims. 
This aligns with the inherently multidisciplinary nature of the HRI field, as highlighted by \cite{bartneck2020human}. Technical aspects were the focus of the field since the raise of HRI  \cite{bartneck2011end}. Consequently, HRI in recent years incorporates insights from various disciplines, including ethics and design, in addition to technical contributions. 

We classified technical studies based on the computational models employed, perception and generation capabilities of the affective robot explored, and the robotic platform used for such applications. 
%\hl{@micol: add here more HRI and RO-MAN references as Minja did}

\subsubsection{Computational Models}
\begin{figure}
    \centering
    \includegraphics[width = \columnwidth]{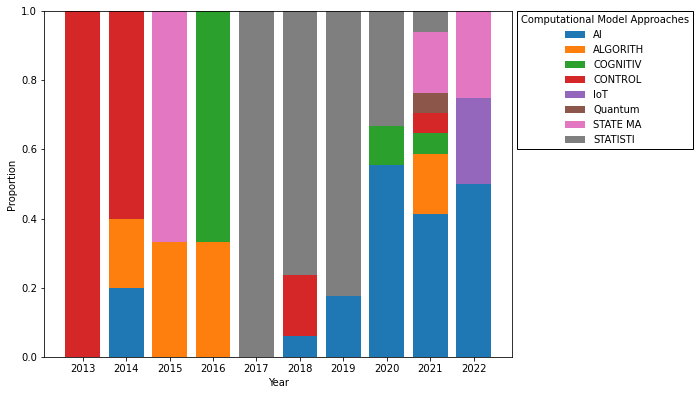}
    \caption{Proportion of computational model approaches per year. "AI" stands for the application of AI models like machine learning, deep learning, and reinforcement learning, "ALGORITH" stands for algorithmic based operation system, "COGNITIV" stands for computational models aimed at following cognitive principles, "CONTROL" stands for human control systems, "IoT" stands for Internet of Things - systems that facilitates communication between devices and the cloud, "Quantum" stands for computational models that are based on quantum computing, "STATE MA" stands for State Machine - a behavior model that consists of a finite number of states, "STATISTI" stands for the application of statistical model for empirical evaluation of affective robotic interactions.}
    \label{fig:cma}
\end{figure}

In terms of computational models employed in affective robots, interestingly, until 2017, studies employed affective robots using variety of computational techniques (e.g., control systems, algorithms, state machines and cognitive architectures) as shown in Figure \ref{fig:cma}. However, between 2017 and 2019, studies started employing statistical models via empirical research, with almost all of the research conducted being empirical. Generally, statistical models have been the most popular computational models used in affective robotics studies (46\% of the studies). This is followed by studies deploying AI-based models on affective robots (27\%) (such as deep learning models, machine learning, reinforcement learning, and natural language processing), which gained popularity in 2020 (more than 50\% of the studies), 2021 (40\% of the studies) and 2022 (50\% of the studies). The increasing use of AI-based models over statistical methods signifies a desire within the field to develop more sophisticated models aspiring for automation.
These models aim to accurately learn from large datasets to recognize, interpret, and respond to human affective states, thus adapting to the state-of-the-art in affective computing and machine learning at large. 

\subsubsection{Affective Capabilities}
The affective perception capabilities embedded in robots evolved from a single modality emotion recognition model, where the majority of the studies used facial expression recognition systems until 2020, to a multi-modal emotion recognition approach in which authors have begun to explore data-driven approaches from multiple data sources (e.g., in \cite{hefter2021development}, the authors considered facial expressions and touch behaviour to detect emotions). 
While autonomous affective perception in studies primarily relied on the robot's ability to perceive emotions from facial expressions without human intervention, a different pattern emerged for affective generation capabilities of robots that utilized both WoZ (Wizard of Oz) and autonomous data-driven computational models. This is due to the need of controling the interaction to ensure most of the effects come from the generated behaviour of the robot, and recently this trend is evolving by using more autonomous generation. This shift is due to the current advancement in AI in general, and generative AI in particular \cite{sakirin2023survey} that exploded in the last few years. However, the generation of human-like behaviours (e.g., facial, speech, gestures) is still a challenge that needs further investigation within affective computing \cite{kucherenko2023genea}. 

Most studies on behavior generation in human-robot interaction have focused on linking generation to a specific interaction paradigm, rather than solely investigating computational affect generation models. For example,\cite{javed2019interactions} designed a robot that could generate empathic behaviors in order to optimize user engagement. This approach prioritizes creating interactions that foster empathy, as opposed to solely exploring computational models of affect generation.
% (e.g.,  designed a robot that could generate empathic behaviour to optimise users' engagement) rather than the investigation of solely computational affective generation models in human-robot interaction. This could be because .. \hl{add a discussion here on this aspect.}
The development and implementation of computational models to generate affective behaviours may be very challenging and it may also have limitations in capturing the nuances and contextual factors involved in human emotional expression and perception. 
Therefore, grounding affect generation in specific interaction paradigms like human-robot interaction in well-being applications may yield more insightful results for better understanding the complex nature of affective behaviours.

\subsubsection{Robotic Platforms}
Lastly, we identified humanoid robots as the most common robotic platforms (52\%) within the technical studies. They are followed by machine-like (17\%), pet-like (11\%), and bio-inspired (2\%) robots. The remaining studies did not report the robotic platform used. 
%Within the last decade, we observed that the humanoid robotic platforms used (e.g., Nao, Pepper etc.) have not changed, but the spread use of computational models has increased largely very recently. 
%This might raise several questions related to the appropriateness of such robotic platform for AI-based interactions.
Over the past decade, humanoid robotic platforms like Nao and Pepper have remained relatively unchanged, while the adoption of computational models has increased significantly. This raises concerns about the appropriateness of these platforms for AI-based interactions, given their limitations in terms of computational capabilities and adaptability.
These platforms, indeed, present to date several limitations, making it more challenging to deploy such robots in everyday lives and for daily applications. 
First, the current robotic platforms lack sufficient local computational power, which is crucial for processing complex AI models and handling large amounts of data in real-time \cite{stasse2019overview}. To date, researchers use largely cloud computing or external service APIs that lead to latency and reduced responsiveness, making interactions less seamless and less natural \cite{spitale2023vita}.
Second, the computational models and algorithms used in these robotic platforms are often not designed to handle the complexity and variability of real-world interactions and to be embedded into a robot. This can result in robots that are not able to effectively learn from their interactions with humans or adapt to contextual situations.
Last, the robotic platforms are often designed for specific tasks (e.g., QT robot for interacting with children with autism \cite{costa2017socially}) or applications, which limits their scalability and flexibility. This makes it difficult to re-use them for different tasks in real-world scenarios.
%\hl{add also the importance to have local computational power.}
%\hl{add limitations of the robotic platform from a computational perspective.}

%Robotics platform limitations? Can we have a systematic analysis of the robotic platform from a computational perspective?

\subsection{Design Research Changes Over Time}

%Design: 
%\begin{itemize}
%    \item RQ.D1: What design methods are used in HRI? (here, methods = Tools, techniques and approaches)
%    \item RQ.D2: Which people are included as users and stakeholders? Who are robots designed with/for? (Socioeconomic status, ethical issues?)
 %   \item RQ.D3: How involved is the user and other stakeholders in current design?
%    \item RQ.D4: How have design methods evolved over time in HRI?
%    \item RQ.D5: How do design methods converge and diverge in different application areas? (e.g. hospital study vs. robot for workplace)
%\end{itemize}

%Look specifically into cross-section of design + affective, and design + ethical

\begin{figure}
    \centering
    \includegraphics[width = \columnwidth]{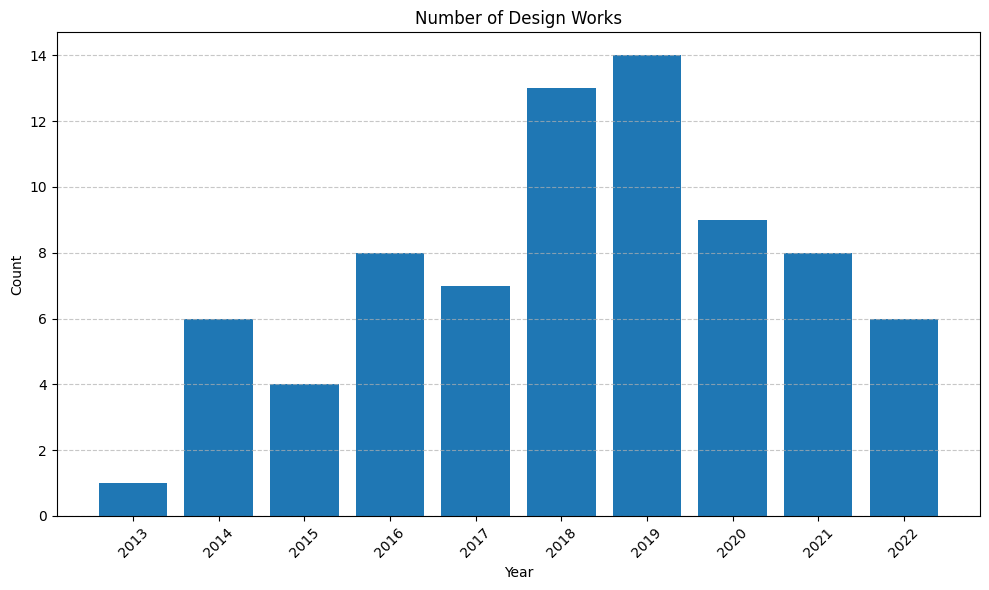}
    \caption{Number of design works by year.}
    \label{fig:design_evolution}
\end{figure}

76 of the surveyed works were classified as having a design focus. In general, our data shows that the most design-focused works in affective robotics were published in 2018 (13 works) and 2019 (14 works), increasing steadily from 2013 (1 work) to this point, and then decreasing from 2020 (9 works) onward (see Fig. \ref{fig:design_evolution}). As pointed out by Lupetti et al. \cite{lupetti2021designerly}, the first explicit design track was introduced in the HRI conference in 2015 \cite{hri2015}, and Ro-Man conference in 2016 \cite{roman2016}. The increasing number of design works in our data aligns with the established design tracks in 2015 and 2016, placing an emphasis on design work published in the following years. 

\begin{figure}
    \centering
    \includegraphics[width = \columnwidth]{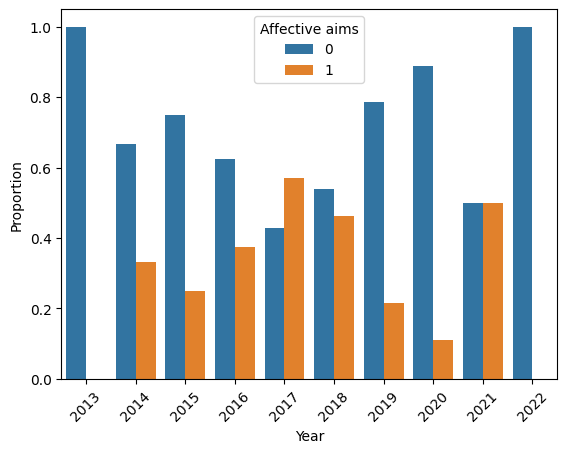}
    \caption{Proportion of affective aims in design studies (1) and affect-related studies (0) by year.}
    \label{fig:design_affective}
\end{figure}

In our data analysis, we classified each paper to have either affective aims (i.e., the study had an explicit aim of exploring the affective design, capability, etc. of a robot) or an affective component (i.e., the paper explores affective robots, but does not have it as its main focus). Out of 76 design works, 24 had affective aims. The peak of the proportion of design studies with affective aims was in 2017 (57\%), where it rose to steadily from 25\% in 2015, and fell steadily to 11\% in 2020 (see Fig. \ref{fig:design_affective}). A recent peak in affective aims was in 2020 with 50\%. However, no design works with affective aims were observed in 2022. These trends may indicate a cyclical interest in exploring affective aims in design studies, with no clear increasing or decreasing trend.

%\subsubsection{RQ.D3: How involved is the user and other stakeholders in current design?}

\subsubsection{Level of User Involvement} 

We classified the 76 identified design studies into categories based on the level of user involvement: 14 participatory design (or co-design), 57 user-centred, and 5 none reported.  %As the field of HRI has not yet agreed on differing definitions of participatory and co-design, we decided to use the same label for both.

\begin{figure}
    \centering
    \includegraphics[width = \columnwidth]{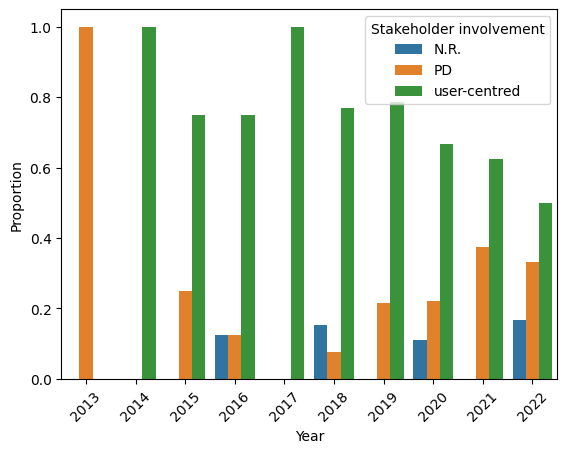}
    \caption{Stakeholder involvement proportions in design studies by year.}
    \label{fig:design_user_proportions}
\end{figure}

%"Consultation participants are informally asked for their opinions, perspectives, ideas and concerns. "Involvement provides more opportunity for input into decision-making, but not agenda setting.Participation allows for more input at the design stages, including defining the agenda or the design activities.Co-production, participants are treated as equals and have an equal say in deciding the project goals and outcomes." https://uxdesign.cc/difference-between-co-design-participatory-design-df4376666816
%participation and co production --> into PD bin
%consultation and involvement --> into UCD bin
%none of these mentioned --> N.R.
% We recognize that this categorization of papers does not take into account the eventuality of involving stakeholders...
%We do not take into further account the level of further stakeholder involvement, e.g. according to \cite{fischer2020importance} ... as this was not relevant to the subject of this review. We invite future researchers to further dig into this aspect and consider e.g. the framework established by ... to examine this.
From 2018 onward, the proportion of reported participatory design contributions has been increasing (8\% in 2018, reaching 38\% in 2021 and 33\% in 2022, see Fig. \ref{fig:design_user_proportions}). According to Bodker, the second wave of HCI which initiated around 1999 and continuing until 2006, placed an emphasis on user-centred design and participatory design \cite{bodker2015third}. As the design tracks at the HRI and Ro-Man conferences were only established in 2018 and 2019, the HRI design trends may be following HCI, with a delayed increase in participatory design when compared to HCI. Indeed, Lupetti et al. \cite{lupetti2021designerly} identify an increasing trend toward user-centred design and including users in the design process through participatory methods in HRI \cite{lupetti2021designerly}. This aligns with the work of Alves-Oliveira et al. \cite{alves2022connecting}, calling for more user-involved design to truly address user needs and problems. 

However, published design works in affective robots for well-being have decreased from 2019 (14 works) to 2022 (6 works). This decrease may be related to the challenges in designing social robots, e.g., design for purpose and artificial emotion expression \cite{alves2022connecting}. Designing affective robots for well-being specifically has several unique challenges, such as appropriately personalizing verbal expressions while preserving well-being practice \cite{axelsson2022robots}, and incorporating stakeholders such as carers into the robot design \cite{moharana2019robots}.  Additional challenges emerge when researchers attempt to design affective robots intended to be deployed in healthcare systems for direct patient care. Designs must consider how an affective robot could fit into an already-complex clinical workflow, how to ensure appropriate patient privacy and risk management (e.g., in the case of suicidal thoughts expressed), or how to prioritize stakeholder engagement with a multidisciplinary treatment team. Such factors may make the (co-)design of affective robots for well-being more challenging.

%Why are the design studies decreasing?

%Proportion of PD is increasing from 2018 - maybe we can tie this to a few papers? user-centred is decerasing but that is to make room for PD. less funding for design in general, but what funding there is is more toward PD?

%\hl{Affective + design cross-section discussion? Maybe we should systematically decide how to approach this in the paper - or are the necessary figures covered in the starting section?}

\subsubsection{Who Are Involved in Design Studies} 

In the PD studies, 6 out of the 14 studies reported children or teenagers as co-designers. The WHO has recommended early detection and intervention as one of the key strategies in improving young people's mental health and resilience \cite{WHO_mental-health}. Using PD strategies with this group in particular may reflect one of the key aims of PD: empowering the users of the designed technology \cite{ertner2010five}. Three of the PD studies described involving mental health professionals (e.g. psychologists or mental health coaches), and two mentioned involving familial caregivers. %\hl{Only one paper did not involve the target user group. maybe check this is it even worth including?}

In the 57 user-centred studies, 22 involved clinicians (e.g., healthcare staff or disability service workers), and 5 of them involved family members or informal caregivers in the design process (four of these studies focused on older adults and their unique challenges (e.g., dementia), and one for children's kinesiology. 29 of the studies explicitly collaborated with (prospective) end-users. %\hl{maybe this should be checked since it's a big claim - I have counted them by hand but not double checked them and the data is a bit messy}. 
These figures reflect that most of the studies focused on collaborating with only one of the stakeholder groups (i.e. end-users or other stakeholders). Only 9 of the studies focused on multiple stakeholder groups (e.g., both patients and clinicians). The tendency to focus on a single stakeholder group is understandable from a pragmatic perspective but may limit external validity given the highly multidisciplinary nature of applied patient care. Given that many affective robots are intended to be deployed in healthcare settings, an important future direction for user-centered design studies will be to involve patients, caregivers, and a range of different provider types, ideally together as part of a collaborative design process. %\hl{I think this is surprising - should we try to create some kind of recommendation about including multiple stakeholders in the same room, or is that premature? maybe we could discuss the pros and cons of this? Cate, opinions? should patients be involved in the design of this technology and does it depend e.g. on clinical severity? do they even want to be and is it fair to "push that on them"? in which cases would it be especially important? Also something to consider: studies that focus on one stakeholder group only may be part of a larger design effort involving multiple stakeholders.} 

\subsubsection{Ethics in Design Research}

\begin{figure}
    \centering
    \includegraphics[width = \columnwidth]{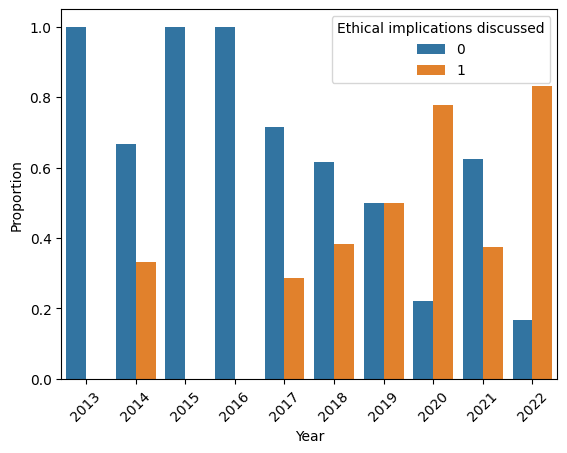}
    \caption{Proportion of design works that discussed ethical implications by year.}
    \label{fig:design_ethics}
\end{figure}

The number of design studies that discuss ethical implications has been generally increasing, climbing from 29\% in 2017 to 83\% in 2022 (see Fig. \ref{fig:design_ethics}). In recent years, works have called for the integration of ethics into the design process \cite{axelsson2022robots}, as well as striving toward equitable robot design  \cite{ostrowski2022ethics}. These works emphasize the importance of diverse user and stakeholder involvement \cite{axelsson2021social, ostrowski2022ethics}. Overall in our dataset (from 2013-2022), a higher percentage of design studies with \textit{vulnerable} user groups (e.g., neurodivergent children, people with PTSD or other mental health issues) discussed ethical implications (43\%), compared to studies with non-vulnerable user groups (31\%). This is not surprising and encouraging as it is often suggested that vulnerable populations can be disproportionately impacted by unethical design choices \cite{ong21}. 

%\begin{figure}
%    \centering
%    \includegraphics[width = \columnwidth]{design_stakeholder_number.png}
%    \caption{How many design studies of each user-involvement category per year.}
    %\label{fig:design_user_per-year}
%\end{figure}

\subsection{Ethical Considerations and Guidelines}

%For each paper we noted: 1. if the paper had a primary ethical theme, 2. if ethical implications of the work were discussed in any capacity, and 3. whether the authors disclosed ethics approval for their research. 
Overall, very few papers were written with a dedicated critique of the ethical concerns related to their work, e.g. discussing safety, fairness, privacy etc. of their application. Only 36 papers were identified as having a primarily \emph{ethical} theme (i.e., the primary contribution is a discussion of the ethics of robots for well-being), and only 41\% of papers had any mention of ethical implications. As such it is difficult to complete a quantitative assessment of ethical implications. We will instead perform a more qualitative assessment, classifying discussions and discussing trends.

\subsubsection{Discussions of Ethical Implications}

\begin{figure}
    \centering
    \includegraphics[width = \columnwidth]{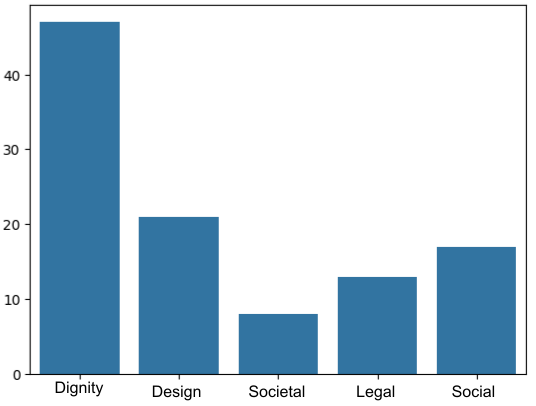}
    \caption{Counts of ethics categories.}
    \label{fig:ethics-categories}
\end{figure}

For each of the papers that was flagged to have discussed ethics in some form a freehand note was provided with the keywords and main topics of concern brought forth in the paper. Multiple frameworks exist to guide ethical discussions of affective systems including \cite{ong21},\cite{chiang21} however, these systems have a primarily AI focus. We had only 40\% of our papers using autonomous capabilities \cite{chiang21} with 14\% having affect generation, less than 1\% having affect perception \cite{ong21}, and 16\% having both affect generation and perception capabilities. However, as part of our inclusion criteria, all papers were required to include a social robot. As such, to guide the conversation on ethical implications we will refer to ``A Code of Ethics for the Human-Robot Interaction Profession" \cite{Riek2014ACO}. Here the authors suggest four primary categories of design principles to consider for human interaction:  Human dignity considerations, Design considerations, Legal considerations, and Social considerations. Each of the noted discussion topics were organized into one of these 4 categories with one additional category ``societal considerations" that includes topics pertaining to infrastructure, employment and similar topics. In Fig. \ref{fig:ethics-categories} we see that dignity and design implications are most often discussed. Examples of dignity considerations that were seen include emotional needs, such as: autonomy, person-hood, and preference; rights to privacy, such as: monitoring, data privacy and policing; and respect for humans' emotional and physical frailty, such as: negative reactions, emotional exploitation, embarrassment, infantilization, powerlessness, infection, and physical safety. For design consideration, we saw commentary on topics such as, trust, reliability, equity, fairness, ableism, control, transparency, accessibility and maleficence. Legal considerations included topics such as informed consent, liability and accountability, and social considerations included attachment, deception and coercion.

\begin{figure}
    \centering
    \includegraphics[width = \columnwidth]{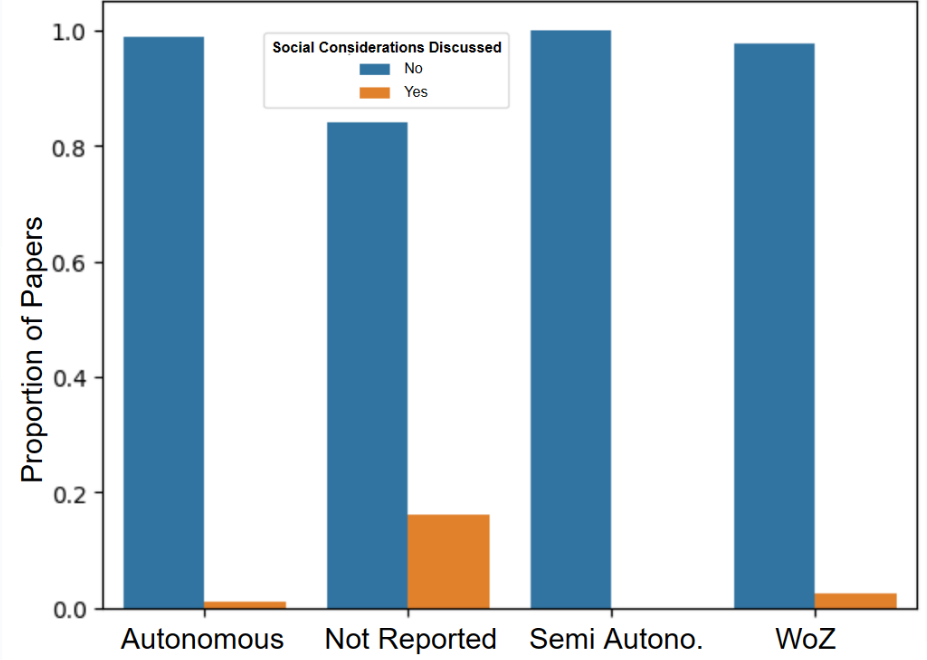}
    \caption{Proportion of papers with social discussion topics by robot operation.}
    \label{fig:ethics-woz}
\end{figure}

Although we see a vast array of ethical topics discussed, we note that instead of many papers each commenting on a few ethical considerations impacting their study, we instead saw papers (69) that had more thorough discussions of several ethical implications. Regarding the sub consideration provided by Riek and Howard \cite{Riek2014ACO}, all of the Human Dignity considerations are considered in ethical discussions of our sample, however, the other three considerations are missing several applicable discussion points. We instead see the authors opting to consistently refer to the same topics presented in the previous paragraph. For design considerations there is a lack of consideration of opt-outs and kill switches, real-time status indicators and predictability of behaviour. Legal considerations, in general, are not thoroughly discussed. Yet, with the advent of the new EU Artificial Intelligence Act \cite{eu-ai-act} these discussions will become exceedingly important. For social considerations, two important considerations were lacking: reducing the use of WoZ to avoid Turing deceptions \cite{Riek2010TheAO, miller10} and limiting humanoid morphology. Both of these techniques are regularly employed in social robotics for well-being with 40\% of papers using WoZ techniques and 52\% using humanoid robots. However, we do not see an increase in the discussion of social considerations in WoZ studies (Fig. \ref{fig:ethics-woz}). Roboticists will often employ WoZ to pilot ideas and reduce the burden of programming as well as the possibly of mistakes, however, these techniques involve deceit and can be particularly difficult for vulnerable uses and may result in increased expectations for robot behaviour that contradicts the considerations of predictability in design. 

%\subsubsection{Trends in Ethics Inclusion}

\subsubsection{Disclosure of Ethics Approval} 
\begin{figure*}
    \centering
    \includegraphics[width = \textwidth]{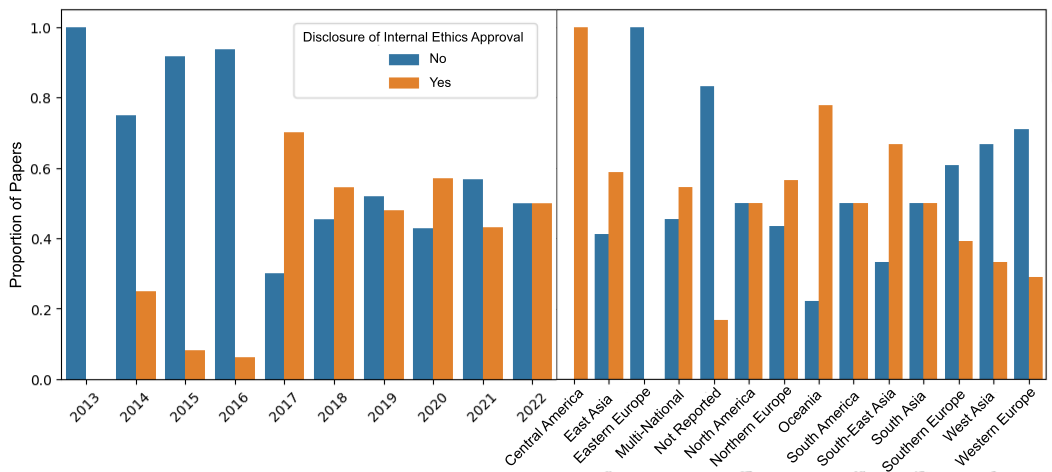}
    \caption{Proportion of papers providing internal ethics disclosure, left: by year, right: by geographic region}
    \label{fig:ethics-internal-disclosure}
\end{figure*}

With the recent focus on Equity, Diversity and Inclusion (EDI), the number of institutions and granting agencies requiring a disclosure of ethics approval appears to be increasing. External approval shows no trends, with only 10 total papers (5\%) disclosing external approval. However, we do see the expected trend in internal ethics approval disclosure. Overall 101 (47\%) of the papers disclosed internal approval, yet, as can be seen in Fig. \ref{fig:ethics-internal-disclosure} (left), previous to 2017 the proportion ethics approval disclosures was quite low. Although there has been an increase, we are still only seeing disclosure of ethics approvals in about half of the papers. These results tell us that most institutions are handling ethics themselves. Although it appears only half of the research institutions are requiring a disclosure of ethics approval, it is, nevertheless, difficult to comment on the amount of institutions requiring ethics approval for these studies as this amount is likely higher than that of the disclosures as researchers themselves may have chosen not to disclose the approval. Geographically, we see that Eastern, Western, and Southern Europe, as well as Western Asia stand out as having a lower ethics disclosure proportions, Fig. \ref{fig:ethics-internal-disclosure} (right). These discrepancies may be a result of funding and institutional regulations where the institution or funder requires the disclosure of ethics approval, or possibly cultural norms \cite{Antes2018-ANTTRO-12}.

\subsubsection{Application Scenarios and Populations} 
\begin{figure}
    \centering
    \includegraphics[width = \columnwidth]{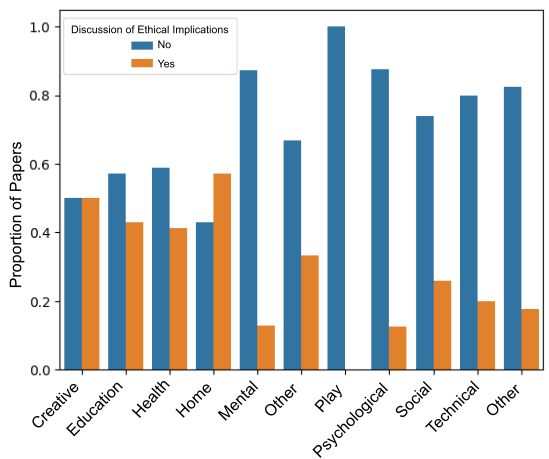}
    \caption{Proportion of papers discussing ethical implications by application scenario.}
    \label{fig:ethics-app-scenario}
\end{figure}

\begin{figure}
    \centering
    \includegraphics[width = \columnwidth]{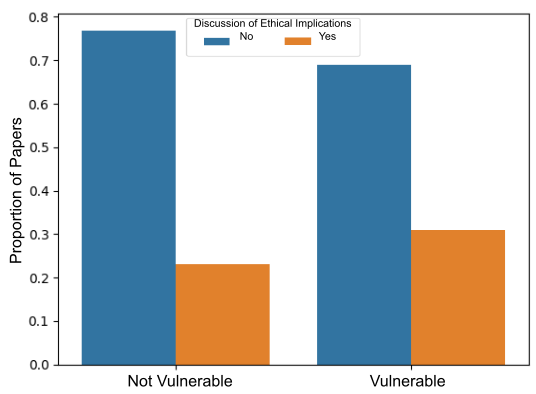}
    \caption{Proportion of papers discussing ethical implications for vulnerable populations.}
    \label{fig:ethics-vulnerable}
\end{figure}

Observing the proportions, interestingly, the highest proportion of papers containing ethical discussion are in applications within the home (Fig. \ref{fig:ethics-app-scenario}), followed by those employed to increase children's creativity, education, and health. Most of these scenarios likely include interaction with children, which may play a role in the need to carefully consider ethical implications. Interestingly, psychological use cases have a lower proportion of ethics discussions, despite the sensitive nature of these applications. To delve further into this research question, we explored whether research where the target population was primarily vulnerable (e.g., neurodivergent children, people with PTSD or other mental health issues) more often discussed ethical considerations \cite{ptsd2022}. Vulnerable target populations included, but were not limited to those involving children, the elderly and people with intellectual and physical disabilities. Yet, we do not see a noticeable difference in the proportion of papers discussing ethics for vulnerable populations (Fig. \ref{fig:ethics-vulnerable}). 

\begin{figure}
    \centering
    \includegraphics[width = \columnwidth]{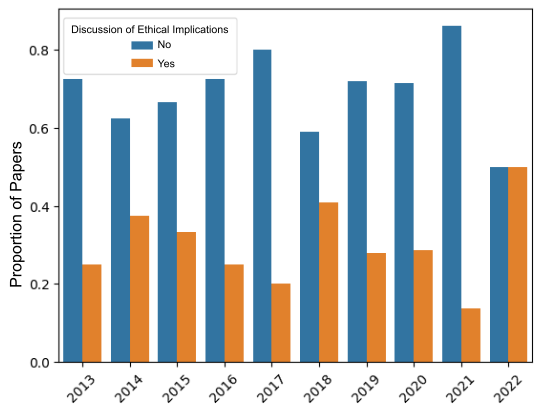}
    \caption{Proportion of papers discussing ethical implications by year.}
    \label{fig:ethics-yearly}
\end{figure}

Unlike with the disclosure of ethics approval we do not see a trend in the discussion of ethical implications over time, Fig. \ref{fig:ethics-yearly}. The sole outlier here is the most recent year, 2022, where the number of papers discussing ethical implications was equal to those that did not. However, it is not possible to speculate if this was unique to this year of publication, or whether it will continue to the future.

%\subsection{Affect Recognition and Generation}
%\section{Discussion}

\section{10 Years of Affective Robotics}
This section reports the most relevant results on research methodology, topic, aims, outcomes, and robotic operations collected from our systematic review by analysing the last decade in affective robotics for well-being, addressing \textbf{C3}. 

Note that a comprehensive report of the results can be found in the Supplementary Material, and that this section is only meant to give the reader an overview of affective robotics for well-being field to better understand the evolution of the technological, design, and ethical aspects of the field in the last decade, as described in Section \ref{sec:evolution}.

\subsection{Research Methodology}
Our observations indicate high variability in the research methodology of affective robotics studies, marked by variations in  data collection methods, participant recruitment, and the duration of studies. 

\subsubsection{Data Collection methods and type of studies} 
\begin{figure}[h!]
    \centering
    \includegraphics[width = \columnwidth]{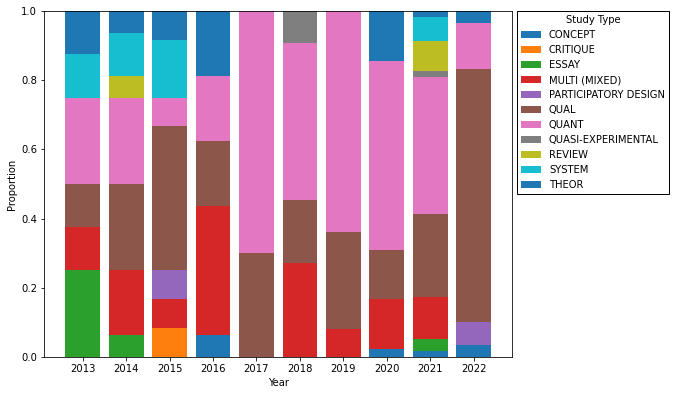}
    \caption{Proportion of study types per each year.}
    \label{fig:st}
\end{figure}

We observed interesting trends in data collection methods and their evolution over the years as shown in Figures \ref{fig:st}. Quantitative studies have predominantly shaped the field of affective robotics, accounting for 39\% of the studies published and showcasing a methodological preference for measurable, data-driven insights. Although quantitative studies predominate  the field, qualitative and mixed-methods studies are also prevalent. Qualitative studies account for 28.9\% of the studies, and mixed-methods account for 13.4\% of the studies. Over time, we have observed a notable increase in the number of qualitative and mixed-methods studies, indicating a gradual shift towards a more holistic understanding of affective interactions with social robots. Specifically, there was a marked increase in qualitative studies in 2022 (75\% of the studies). This could be due to restrictions on conducting behavioural experiments during the COVID-19 pandemic, leading many researchers to work with smaller samples and rely on qualitative methods. This evolving blend of methodologies highlights a dynamic, multifaceted research landscape that adapts to the complex nature of affective interactions in robotics. %Interestingly, participatory design studies are substantially underrepresented (1\% of the total number of studies, 9\% of the studies in 2015 and 7\% of the studies in 2022), despite their potential to significantly contribute to affective robotics by emphasizing user engagement in development processes. 

Another point of interest is the low proportion of system studies (4\% of the total number of studies), with most studies published between 2013 (13\% of the studies) and 2015 (17\%). It is presumed that during this period, the field of affective robotics was in a more developmental stage, focusing on building and assessing foundational systems crucial for establishing baselines and understanding the capabilities of these affective agents. Post-2015, the focus might have shifted towards refining these systems based on initial findings and integrating them into more complex user studies to explore the broader implications of affective robotics in various contexts and settings. This shift could be an indicator of technological maturity, where incremental improvements and applications of existing systems became more relevant for advancing the field. For example, as affect detection models were advancing in the broader affective computing community, affective robotics researchers gradually applied these advancements empirically to their own systems rather than developing dedicated systems from scratch \cite{Kappas2023}. Another possibility is that system-related research in affective robotics is now broader, encompassing disciplines like social robotics, affective computing, machine learning, natural language processing, and computer vision \cite{Afzal2023,Wang2022}, with these systems then applied in affective robotics research.

\subsubsection{Number of Participants} 
\begin{figure}
    \centering
    \includegraphics[width = \columnwidth]{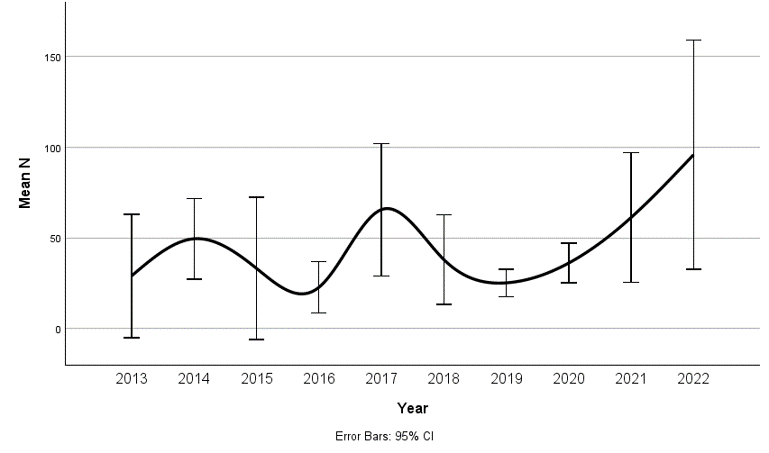}
    \caption{Average number of participants in a study per year}
    \label{fig:np}
\end{figure}

We observed growth in the average number of participants per study over the years, indicating a trend towards larger sample sizes to enhance the robustness of findings as shown in Figure \ref{fig:np}. This increase, from approximately 30 participants per study on average in 2013 to almost 100 participants per study on average by 2022, underscores efforts to ensure findings are robust and widely applicable, essential for technologies that closely interact with human emotions and behaviours. It should be noted that some of these studies are neither empirical nor quantitative in nature, and thus the extent of their contribution and methodological rigour is not assessed by the size of their sample.

%To further examine this observation, we conducted a simple regression analysis finding a small yet significant effect, $\beta$ = 5.32, \textit{STD} = 2.61, $b*$ = .15, $p$ = .043, 95\%CI [.17,10.48], $R^2$ = 2\% (.021). Nevertheless, it is important to consider that since the effect size is small, the trend is inconsistent. 

\subsubsection{Study Sessions} 
\begin{figure}
    \centering
    \includegraphics[width = \columnwidth]{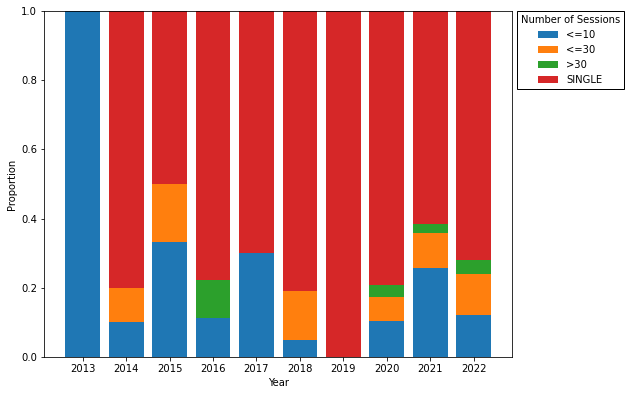}
    \caption{Proportion of studies with employing different number of sessions per each year.}
    \label{fig:ss}
\end{figure}

We found that single-session studies have emerged as the predominant focus in affective robotics (72\% of the total number of studies), with a notable peak in long-term studies (lasting up to 10 sessions) published in 2021 (39\% of the studies), as depicted in Figure \ref{fig:ss}. However, the overall trend for long-term studies is characterized by sporadic spikes rather than a consistent increase or decrease. This preference for single-session studies may be attributed to researchers prioritizing the assessment of technological developments and the introduction of new affective features. Such studies allow for rapid validation of innovations, keeping pace with the swift technological advancements in the field \cite{spitale2022affective}. This approach underscores a dynamic and evolving research ecosystem, where the drive for innovation often outweighs the desire for long-term deployment insights. The fast-paced nature of technological progress in affective robotics may also explain this preference for single-session studies: long-term studies could be perceived by researchers as less relevant, as technology could become outdated by the time a study concludes \cite{Haefner2021}. Additionally, logistical and financial constraints, along with the challenges of maintaining participant engagement over extended periods, may discourage longer study durations \cite{Heckert2020,blt_2023}.

\subsection{Aims and Applications}

Affective robotics is a relatively wide area of research, encompassing studies with various aims, disciplinary affiliations, and application scenarios.

\begin{figure*}
    \centering
    \includegraphics[width =.32 \textwidth]{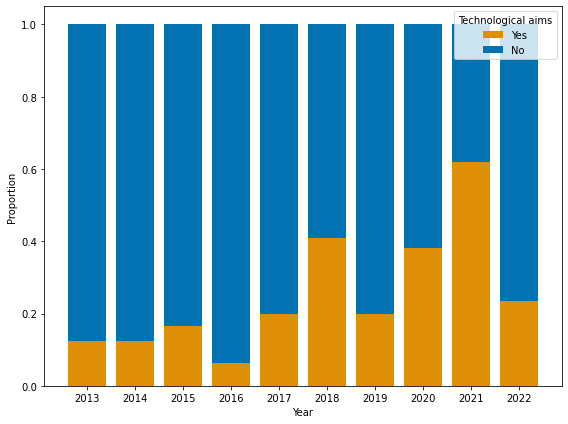}
    \includegraphics[width =.32 \textwidth]{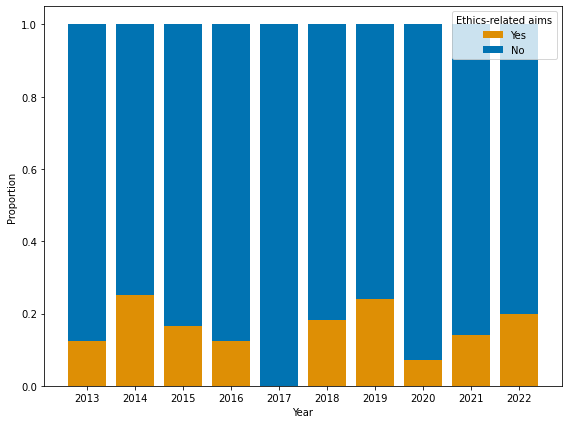}
    \includegraphics[width =.32 \textwidth]{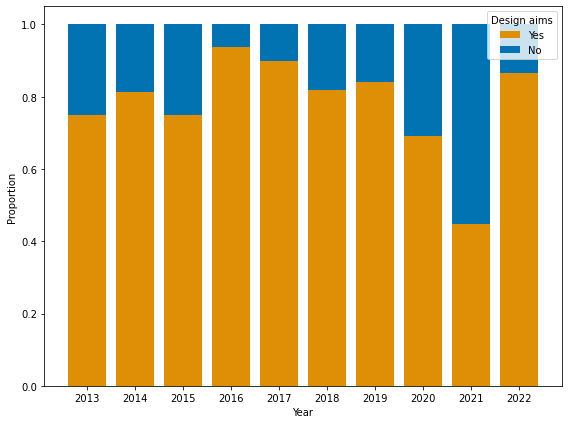}
    \caption{From top left to right: (1) proportion of studies including technological aims per year. (2) proportion of studies including ethical aims per year. (3) proportion of studies including design aims per year. }
    \label{fig:aims}
\end{figure*}

\subsubsection{Aims}
Three main aims were identified in the literature published between 2013 and 2022: technological, ethical, and design, as shown in Figure \ref{fig:aims}. 
Regarding technological aims, we observed a steady positive trend in the proportion of studies including technological aims over the years, reaching a peak of over 60\% of studies published with technological aims in 2021. %We further validated this trend with binary logistic regression, revealing a significant 27\% increase in the odds of a study incorporating a technological aim for each passing year. 
In terms of ethical aims, our data show that studies published rarely include these aims, with a maximum of slightly more than 20\% of the studies including ethical aims (in 2014, 2019, and 2022). 
Regarding design aims, most studies published in affective robotics tend to include such aims, ranging between 70\% to 95\% of the studies published per year, with the exception of 45\% in 2021. %Nevertheless, further examination using binary logistic regression suggests a decline in the likelihood of a study consisting design-related aims as the years advance. More precisely, we observed an approximate 14\% decrease in the odds of a study having a design-related aim with each passing year. 
%Finally, in terms of affective aims, while we observed a steady positive trend in the research published that includes affective aims over the years, these aims range only between 10\% to 60\% at most. %This might be due to a combination of factors such as the complexity of accurately modeling and interpreting human emotions and the high costs associated with developing affect-sensitive technologies. Accordingly, it could be (as observed in other aspects of the data), that researchers interested with affective robotics prioritised their exploration via other means (e.g., 

Our results show that only three papers (1\%) cover all three aspects, namely technological, design and ethical, while 14\% focused on technical and design, 2\% on technical and ethical, and the 11\% on ethical and design aspects. The majority of them (72\%) focused on only one aspect. The limited integration of technological, design, and ethical aspects in research could potentially be attributed to the complexities and resource constraints of addressing multiple aims simultaneously, which is a common challenge in human-centred engineering and computer science research \cite{Stephanidis2019,hcd2022}. Specialization in one aspect is often required by the need for depth and clarity in research, especially in early stages, alongside academic and publication pressures that prioritize quicker, focused studies. As the field evolves, interdisciplinary collaborations may facilitate more holistic approaches with varying aims.

%\hl{@GUY: add a brief consideration about these numbers. why people is not taking into account more aspects at the same time? add numbers of single aspect papers.}

\subsubsection{Discipline Affiliations}
The research diversity in affective robotics is also evidenced in the presence of single and multidisciplinary teams, and disciplinary affiliations. Over the years we can observe that there are more studies published by single-discipline teams over multidisciplinary teams, with the exception of 2018 with 61.9\% of the studies published being by multidisciplinary teams. Nonetheless, in most years we can observe that between 27\% (in 2019) to 44\% (in 2015 and 2017) of the studies published in each year were by multidisciplinary teams (with the exception of 6\% of the studies in 2016, and 61.9\% of the studies in 2018). This seems like a relatively high ratio of multidisciplinary efforts. 

\begin{figure}
    \centering
    \includegraphics[width = 0.35\textwidth]{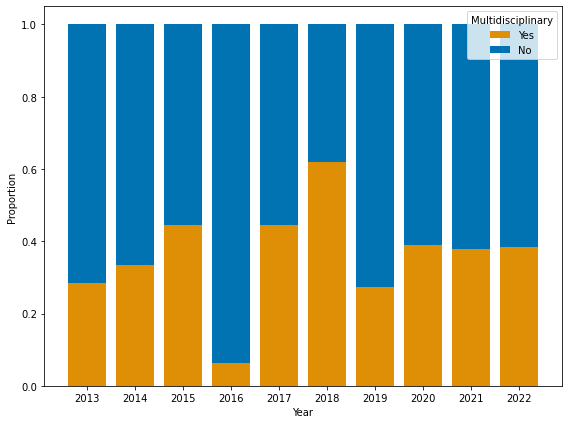}
    \caption{Proportion of studies with multidisciplinary and single discipline teams per each year.}
    \label{fig:msd}
\end{figure}

Single discipline studies have seen a surge of papers from authors with a computer science background, followed by papers with authors with an engineering background, and then a media background. Surprisingly, single discipline studies coming from psychology see only a maximum of 30\% of the papers in a year (in 2018). Given that ‘\textit{Affect}' at its core is a psychological construct \cite{Barrett2009}, and given the broader connection of the field to ‘\textit{Psychological Well-being}' \cite{robohealth2024}, it is expected that researchers with a background in psychology would be more prominently involved, similar to related fields such as social robotics and HRI \cite{Henschel2021}. Studies published by multidisciplinary teams sees similar trends, with most of the first authors in these teams having computer science background. Accordingly, we can assume that despite the multidisciplinary nature and demands of affective robotics research, the field is still primarily driven by technical questions and aims.   

\begin{figure*}
    \centering
    \includegraphics[width = .49\textwidth]{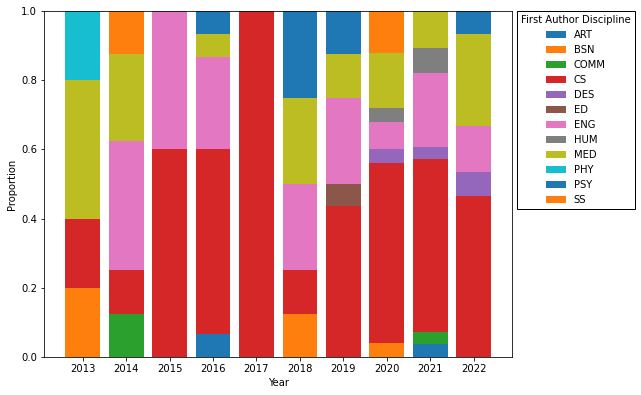}
    \includegraphics[width = .49\textwidth]{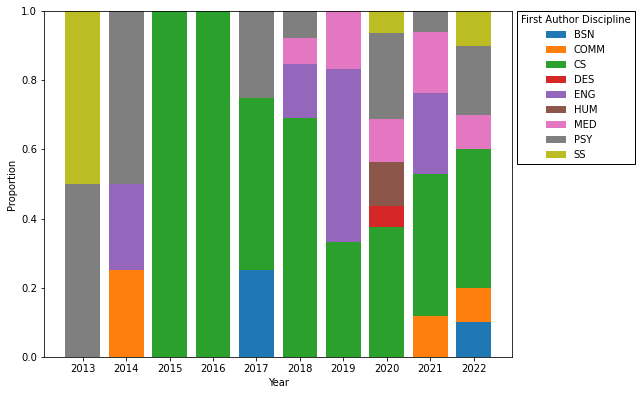}
    \caption{From left to right: (1) Proportion of first author disciplinary affiliation within studies with single discipline teams per each year. (2) Proportion of first author disciplinary affiliation within studies with multidisciplinary teams per each year. }
    \label{fig:msd}
\end{figure*}

\subsubsection{Application Contexts} 
We observed that a substantial proportion of the studies published in affective robotics are concerned with health application scenarios (28\% of all studies) ranging between 20\% (in 2019) to 58\% (in 2015) of the studies published in a year (except for 10\% in 2017). This is followed by studies concerned with social application scenarios (21\% of all studies), ranging between 6\% (in 2016) and 31\% (in 2014) of the studies published in a year (except for 0\% in 2013). This is an important trend in affective robotics, stressing that despite the technical aim that dominates the field (with explicit research aims and disciplinary affiliations), many of the studies are aimed at being applied in typical social contexts that are customary in affective computing (i.e., health and social settings). Following these two application scenarios (i.e., health and social settings), the third most prominent application scenario is mental health, constituting 16.3\% of the papers. Hence, while socially assistive robots have been studied in a variety of health-related settings, including physical rehabilitation and primary care \cite{robohealth2024}, research into affective robotics considers role of social robots in mental health settings and other applied care scenarios to be critical within the field. 

\begin{figure}
    \centering
    \includegraphics[width = \columnwidth]{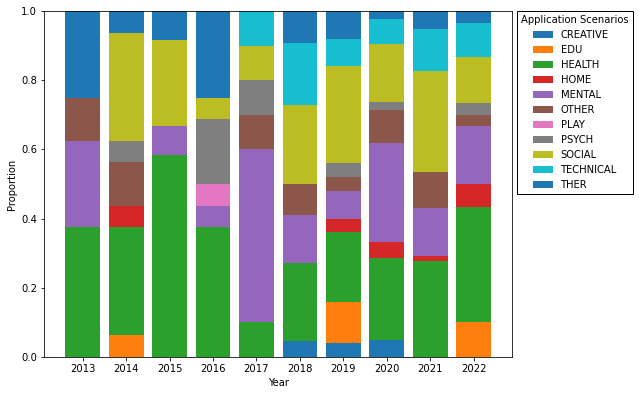}
    \caption{Proportion of studies employed in different application scenarios per each year.}
    \label{fig:aps}
\end{figure}

This approach is also evidenced by the growing proportion of studies focused on well-being target outcomes, increasing from 17\% in 2014 to 36\% in 2022, with a peak of 55\% of all psychological target outcomes in affective robotics research observed in 2020. However, it is important to consider the generality and breadth of the field. This is evidenced from the proportion of studies with multiple outcomes, ranging from 65\% of the psychological target outcomes of affective robotics papers in 2013, to 80\% of the affective robotics papers published in 2016. As noted previously, there has been a shift towards more studies investigating well-being outcomes—a term that is in itself vague and broad. Nonetheless, starting in 2017, we began to observe a diversification of outcomes studied, with many (15 identified) unique psychological target outcomes related to well-being identified, such as eating disorders, eldercare, sleeping habits, stress-related issues, and psychopathologies, among others. This diversification suggests that beyond the growing interest in the health and well-being applications of affective robots, the field is maturing. Researchers are increasingly seeking to assess the applicability of this technology for addressing specific outcomes (e.g., behavioural changes related to eating and sleep, or emotional support for loneliness and stress), rather than attempting to capture multiple outcomes in single studies \cite{robohealth2024}.

\begin{figure}
    \centering
    \includegraphics[width = \columnwidth]{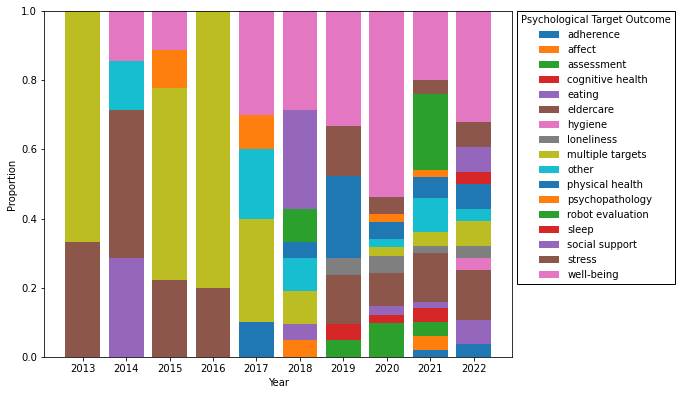}
    \caption{Proportion of studies employed for studying different psychological target outcomes per each year.}
    \label{fig:pto}
\end{figure}

\section{Future Opportunities in Affective Robotics For Well-being}

\begin{table*}[htb!]
\caption{Past, present and future of affective robotics across technical, design, and ethical aspects. }
\footnotesize
\label{tab:evolution}
\resizebox{\textwidth}{!}{%
\begin{tabular}{llll}
\toprule
\textbf{Aim}   & \textbf{Past}                                            & \textbf{Present} & \textbf{Future} \\
\midrule

Technological& Models embedded in robots were& Models are characterised by machine & Research should investigate\\
(Models)& mainly about control systems&  learning and AI-based methods& foundation models for both perception \\
& &  & and generation \\
Technological & Humanoid robotic platforms have & Humanoid robotic plaforms are & New robotic platforms that should enable  \\
(Robot)& been developed with pre-programmed&  the same, and external computational& AI-based models to run \\
& functioning&  locally& \\
\midrule
Design & Increasing interest in designing& Lack of interest in designing& Design works should focus \\
(Affective Interest) &robots with affective purposes& robot for affective purposes& on the human-centered affective capabilities \\
Design &Few works have conducted PD& PD method has been conducted with& Co-design, like PD, works  \\
(User & studies& single stakeholder groups& should include multiple groups\\
Involvement) &&& of stakeholders\\
\midrule
Ethical & Very few works included& About half of studies include & Beyond ethical approval, works should\\
(Inclusion) &information about ethical approval& ethical approval and even fewer& include discussions on implications using  \\
& or implications& include ethical discussion& guidelines like the code of ethics and\\
&&& considering the EU AI Act\\

\bottomrule
\end{tabular}%
}
\end{table*}

Our survey aims at better understanding the evolution of affective robotics for well-being over the last decade. Our results show the past and present of this research field, and this section explores the future opportunities in affective robotics distilled for well-being across technical, design and ethical aspects as collected in Table \ref{tab:evolution}.

Affective robotics is a multi-disciplinary field that includes expertise from computer scientists, psychologists, social scientists and roboticists. As such, this field would benefit from the cross-fertilization of affective computing and robotics \cite{celiktutan2018computational}.
We observed that in the earlier stages of affective computing research, there was more emphasis on empirical, data-driven studies to establish the foundational understanding of affective phenomena \cite{spitale2022affective}. More recently, we note a shift in the research landscape, potentially driven by the very recent technical advancement in the field. This shift in the research approach, from more empirical to potentially more technical, could indicate that the field of affective computing research is maturing and transitioning towards more sophisticated, application-oriented developments.

%\begin{itemize}
%    \item As technology has evolved and the presence of robots has changed over the years, we will be asking how the number and proportion of studies addressing affective aims have changed over time.
%    \item Affective aims can correspond to social and behavioural inquiry, such as studying human-displayed affect or the effect of interaction on human emotion and well-being, as well as to technical and engineering-related inquiries, such as equipping robots with tools and instruments to recognise, simulate, and generate affect. Therefore, we are also interested in examining the disciplinary nature of studies addressing affective aims to better understand how researchers in the community approach these objectives.
%    \item As affective aims can address both behavioural and technical objectives, they can also target a variety of outcomes across different settings. Consequently, we are interested in understanding which psychological outcomes align with these affective aims in research over time, and in identifying the application scenarios in which researchers have investigated these aims throughout the years.
%\end{itemize}

This \textbf{technological} shift in affective robotics for well-being can manifest in different forms, such as advancements in affective sensing and generation techniques, integration of affective AI models into robotic platforms, exploration of real-world applications of these technologies in well-being contexts, and increased interdisciplinary collaboration between technical and social/behavioral researchers. 
As we have already mentioned, the field of affective robotics is increasingly shifting towards data-driven models, and we will observe an even more evident shift in the future driven in part by the emergence of large language models (LLMs) that are revolutionising various domains.
This trend is not limited to affective robotics but is part of a broader movement encompassing human-computer interaction (HCI) \cite{gao2024taxonomy}, computational linguistics \cite{ziems2024can}, and affective computing \cite{liu2024affective}.  It is essential to go beyond solely utilising LLM APIs and consider how these models can be tailored to specific use cases, focusing on quality-centric data rather than quantity, especially for high-stake domains like well-being. 
Improving relationships with the affective computing community and enhancing the benchmarking of models intended for implementation in interactions are crucial steps for advancing the affective robotics field. %It is essential that models are not merely conceptual but are developed with the intention of practical implementation. Affective robotics has the potential to demonstrate the feasibility and innovation of these models in real-world settings as in recent works \cite{mathur2021modeling}. 
Therefore, we encourage researchers to collaborate, leverage current advancements, and explore their practical application in real-life social interactions like well-being with robots.
%\hl{@MICOL: emphasis should be AR for well-being. Missing AR for well-being point of view}

Future investigations in affective robotics for well-being are likely to focus on the technical challenges and opportunities associated with integrating multimodal affective data into LLMs for robotic applications \cite{yang2023contextual, etesam2023emotional}. This could involve exploring novel architectures, training methodologies, and evaluation metrics tailored to the unique requirements of multimodal language understanding in well-being contexts. 
We hope that future researchers can build a new generation of intelligent and emotional robotic systems that can seamlessly process and respond to various sensory inputs, paving the way for enhanced human-robot interaction in well-being.

% the field is gradually leaning towards data-driven models, also due to the advent of llm who is revulatinising the field which is probably the case also for affective robotics. The is evident not only in affective robotics but as part of a greater scheme including hci in general, computational lingustics, affective computing. 
%reproducability, open-source, community, federated models (not only using LLM API as is but thinking how these can be extended towards specific use case , i.e., qualtiy-centric data rather than just quantity of data), open science, pre-registration - crucial for ethics and methodological rigour as well aspecially since our research is aimed at supporting huamans' well being and at times also clinicaly diagnosed indviduals, therefore we need to adhere to the highest standarts of empirical research in the field. 

%Better relationships with the affective computing community, better benchmarking of models to be implemnted in interaction. Eventually models are created to be implemented and not just reported as a proof-of-concept, affective robotics has the potneital to show the feasability of such models and innovation. Accordingly, we would encourge for researchers to collabrate, to adapt current developments and see their application in real-life social interaction with embodied agents like robots. 
From a \textbf{design} point of view, co-design of affective robots for well-being is underexplored, and is a future research opportunity.
Especially in clinical contexts, designing with multiple stakeholders (i.e. clinicians, other healthcare workers, and patients) in the same room could be useful for establishing a dialogue between them, and empowering patients in how future robotic technologies could contribute to their care. This approach is consistent with patient-centered, value-based care \cite{silvola2023co}. 
Given the high barriers to clinician participation in co-design sessions (e.g., scheduling demands; long working hours), brief (1 hour or shorter) web-based participatory design sessions may be preferable to longer in-person sessions \cite{savoy2022electronic}. Such online participatory methods have been proposed, e.g. the Hybrid Robotic Design Model, where design teams work in person at specific points of the design process, and other phases are conducted online \cite{ahtinen2023supporting}. %\hl{MINJA: would you like to expand on this more?}

Regarding \textbf{ethical} considerations, all papers with human participants, should include ethics approval disclosures and discussion. We are seeing a trend towards this, but there is still much room for improvement. Additionally, researchers should familiarize themselves with ethical guidelines. Unfortunately, as of yet, there is no large overarching and agreed upon set of guidelines for affective, social robotic and well-being applications, instead, researchers must asses their own work and use the applicable frameworks such as \cite{Riek2014ACO} for social robotics, \cite{ong21} for perception in affective computing, and \cite{chiang21} for AI in clinical applications, to guide their research and discussions. Ethical implications must be considered at all stages of the study process, and stakeholders and their individual ethical responsibilities must be defined prior to conducting research \cite{ong21}. We saw a focus on ethics on a personal level, i.e. dignity, and a lack of consideration for social and societal implications. As affective technology and robotics becomes increasingly prevalent these considerations are crucial for harmonious, safe, and fair integration of these systems \cite{Jobin2019TheGL}.

Moreover, AI has come under scrutiny and much needed regulations are being put in place \cite{eu-ai-act}. Although these laws regulate industry rather than research applications, researchers should not take this as an opportunity to skirt these rules. Likewise, with the move towards open source research and collaboration, any publicly available models or technologies that can be available for use by enterprises will be required to adhere to the regulations.
    
Emphasizing reproducibility, open-source practices, community collaboration, and the development of federated models is crucial and will enhance adherence to several agreed upon metrics of ethical guidlines including transparency, justice and fairness, non-maleficence, responsibility and privacy \cite{Jobin2019TheGL}, \cite{chiang21}. Practices such as open science and pre-registration are vital for ensuring ethical and methodological rigor, especially given that our research aims to support human well-being, including individuals with clinical diagnoses. Adhering to the highest standards of empirical research in the field is paramount to uphold the integrity and impact of our work.

%In the last year we have seen a growing interest by public and political institutions restulting in poilcies and guidliness for the safe,ethical and responsible implemntation of AI in society (e.g., EU-ACT). iN PARTICULAR, THE AFFECTIVE COMPUTING COMMUNITY HAS BEEN LARGELY EFFECTED by this due to the recent limitations imposed on emotion recogntion methods. Goverments are doing the first attemps to regulate affective computing, nevertheless the moves that has been conducted are rather superficial and not based on evidence. Therefore, affective computing, especially for well being, can provide important evidence showing how the use of affective computing can be implemented in an ethical and responsible way while supporting users. (potentially address some of our insights from the ethics part)
%more ethics research is needed - the current political landscape is calling for more ethics measures, more ethics discussion, and our comunity hasn't been great in discussing it in the past 10 years. We see a room for research and discussion specfically in relation to ethics. 

\section{Conclusion}

This survey presents the evolution of the field of affective robotics for well-being over the last decade. By highlighting the past trends, present challenges, and future opportunities in the field of affective robotics for well-being, this survey aims to guide future researchers in tailoring their work based on the lessons learned and the envisioned trajectory of the field. We encourage researchers to consider the various implications of their work, including technical, design, and ethical considerations, to drive the development of affective robotics towards enhancing human well-being.

% or


% if you want by leaving the argument blan

\section*{Acknowledgment}

M.S. is supported by PNRR-PE-AI FAIR project funded by the NextGeneration EU program. G.L. and H.G. are supported by the EPSRC project ARoEQ under grant ref. EP/R030782/1. M.A. is supported by the Finnish Cultural Foundation. A.L and P.T. are supported by the Rajan Scholar Research Fund, the France Canada Research Fund, NSERC Discovery Grant 06908-2019 and Mitacs Globalink ref. FR103868. %\hl{@ALL: add your fundings}

\ifCLASSOPTIONcaptionsoff
  \newpage
\fi



%\bibliographystyle{IEEEtran}
\bibliographystyle{IEEEtran}
\bibliography{ref}


\begin{IEEEbiographynophoto}{Micol Spitale}
Micol Spitale is an Assistant Professor at the Department of Electronics, Information and Bioengineering at the Politecnico di Milano (Polimi), as well as a Visiting Affiliated Researcher at the University of Cambridge.
In recent years, her research has been focused on the field of Social Robotics, Human-Robot Interaction, and Affective Computing, exploring ways to develop robots that are socio-emotionally adaptive and provide ‘coaching’ to promote wellbeing.
\end{IEEEbiographynophoto}

\begin{IEEEbiographynophoto}{Minja Axelsson} Minja Axelsson is a PhD Student at the Department of Computer Science \& Technology at the University of Cambridge. Her research is focused on the design and ethics of social robots for well-being, as well as users' perceptions of them.

\end{IEEEbiographynophoto}

%\newpage

\begin{IEEEbiographynophoto}{Sooyeon Jeong}
Sooyeon Jeong is an Assistant Professor in the Department of Computer Science at Purdue University. Dr. Jeong designs interactive AI agents to improve people's lives and deploys these agents in-the-wild to evaluate how they can enhance people's wellbeing, health, and learning. 
\end{IEEEbiographynophoto}

\begin{IEEEbiographynophoto}{Paige Tutt\"os\'i}
Paige Tutt\"os\'i is a PhD candidate at Simon Fraser University and a visiting researcher at l’Institut FEMTO-ST, Université Bourgogne-Franche-Comté. Her research is focused on the improvement of robotic voices and ethical implications of human robot interaction.
\end{IEEEbiographynophoto}

\begin{IEEEbiographynophoto}{Caitlin A. Stamatis}
Caitlin A. Stamatis, PhD, is the Chief Clinical Officer at Bruin Health. A clinical psychologist by training, Dr. Stamatis's research focuses on using technology-enabled mental healthcare.
\end{IEEEbiographynophoto}

\begin{IEEEbiographynophoto}{Guy Laban}
Guy Laban is a Postdoctoral Research Associate in the Department of Computer Science \& Technology at the University of Cambridge. His research is aimed at exploring how people communicate their emotions to social robots, and how accordingly these interactions enhance and support emotional well-being.
\end{IEEEbiographynophoto}

\begin{IEEEbiographynophoto}{Angelica Lim}
Dr. Angelica Lim is the Director of the Rosie Lab (\href{www.rosielab.ca}), and an Assistant Professor in the School of Computing Science at Simon Fraser University (SFU). Previously, she led the Emotion and Expressivity teams for the Pepper humanoid robot at SoftBank Robotics. She received her B.Sc. in Computing Science with Artificial Intelligence Specialization from SFU and a Ph.D. and M.Sc. in Computer Science (Intelligence Science) from Kyoto University, Japan. Her research interests include multimodal machine learning, affective computing, and human-robot interaction.
\end{IEEEbiographynophoto}

\begin{IEEEbiographynophoto}{Hatice Gunes}
Hatice Gunes is a Full Professor of Affective Intelligence and Robotics in the Department of Computer Science and Technology, University of Cambridge, and the Director of the \href{https://cambridge-afar.github.io/} {Cambridge AFAR Lab}. She is a former President of the Association for the Advancement of Affective Computing, a former Faculty Fellow of the Alan Turing Institute \-- the UK's national institute for data science and artificial intelligence, and is currently a Fellow of the Engineering and Physical Sciences Research Council UK (EPSRC) and Staff Fellow of Trinity Hall.
\end{IEEEbiographynophoto}

%\vfill


\end{document}